\documentclass{article} 
\usepackage{iclr2027_conference,times}

\usepackage{amsmath,amsfonts,bm}

\def\eqref#1{equation~\ref{#1}}

\def\1{\bm{1}}

\def\vc{{\bm{c}}}

\def\vv{{\bm{v}}}
\def\vw{{\bm{w}}}
\def\vx{{\bm{x}}}

\def\mI{{\bm{I}}}

\DeclareMathAlphabet{\mathsfit}{\encodingdefault}{\sfdefault}{m}{sl}
\SetMathAlphabet{\mathsfit}{bold}{\encodingdefault}{\sfdefault}{bx}{n}

\usepackage{hyperref}
\usepackage{url}

\usepackage{cleveref}
\usepackage{natbib}
\usepackage{xspace}
\usepackage{graphicx}
\usepackage{amssymb}
\usepackage{amsmath}
\usepackage{enumitem}
\usepackage{wrapfig}
\usepackage{multirow}
\usepackage{makecell}
\usepackage{algorithm}
\usepackage{algpseudocode}
\usepackage{booktabs}

\def\ie{\emph{i.e.}}

\definecolor{badscore}{RGB}{178,34,34}   
\definecolor{goodscore}{RGB}{34,139,34}  
\newcommand{\bad}[1]{\textcolor{badscore}{#1}}
\newcommand{\good}[1]{\textcolor{goodscore}{#1}}

\title{CAST: Causal Advantage-Structured Training with Spatially Grounded Compositional Rewards for Diffusion Models}

\author{Shu Yu$^{1,2,3}$ \quad Chaochao Lu$^{1}$\thanks{Corresponding author.}\\
$^{1}$Shanghai Artificial Intelligence Laboratory, Shanghai, China\\
$^{2}$Shanghai Innovation Institute, Shanghai, China\\
$^{3}$Fudan University, Shanghai, China\\
\texttt{\{yushu, luchaochao\}@pjlab.org.cn}}

\iclrfinalcopy
\begin{document}

\maketitle

\begin{abstract}
    Online reinforcement learning has been successfully extended to flow matching for diffusion model (DM) image generation.
    However, this paradigm faces three limitations:
    (1) \textbf{Window selection.} Existing methods typically manually set the stochastic differential equation (SDE) sampling window, \ie, the denoising steps where exploration noise is injected. We instead determine it from each model's denoising trajectory.
    (2) \textbf{Reward saturation.} Current methods rely heavily on scoring models trained on human annotations; we find that such scores are extremely high and nearly indistinguishable on the latest SOTA open-source DMs, making advantage estimation largely ineffective.
    (3) \textbf{Sample inefficiency.} A single scalar reward collapses different failure modes into almost identical scores, leaving minimal gradient guidance for targeted improvement.
    To address these issues, we propose \textbf{CAST} (\textbf{C}ausal \textbf{A}dvantage-\textbf{S}tructured \textbf{T}raining), an RL fine-tuning method for pretrained DMs, which (1) identifies the denoising step at which each model fixes the objects and their spatial arrangement in the image and uses that timing to set the SDE window, (2) decomposes each prompt via Causal Scene Graphs (CSG) into \emph{verifiable-atoms}, \ie, minimal semantic units such as an object, count, attribute, or spatial relation that can each be checked independently, and rewards each atom separately, and (3) projects the signed atom-level advantages into pixel space through teacher-forced attention and uses them to spatially weight the SDE policy objective. We fine-tune two of the strongest open-source DMs, FLUX.2-dev and Qwen-Image-2512, with CAST, and evaluate them on GenEval~2, a compositional benchmark on which these models still fail frequently, and on Qwen-Image-Bench for overall quality. Within almost the same training budget, CAST's improvement over the base model on the most challenging GenEval~2 prompts is up to $3.07\times$ that of Flow-GRPO, while overall generation quality also improves. Our project page is at \url{https://opencausalab.github.io/CAST}.
\end{abstract}
    
\section{Introduction}\label{sec:introduction}

Online reinforcement learning (RL) has proven highly effective at enhancing flow matching for diffusion model (DM) image generation.
Flow-GRPO \citep{liu2025flowgrpo} converts the deterministic ordinary differential equation (ODE) denoising process into a stochastic differential equation (SDE) with matched marginal distributions and applies online policy gradient optimization, substantially improving image aesthetic quality.
Recent extensions \citep{savani2026stepwise, deng2026densegrpo, tong2026alleviating} further introduce per-step rewards, enabling credit assignment along the temporal axis.
These advances show that online RL can substantially improve quality when the optimization signal \citep{kirstain2023pick, wu2023human} is reliable.

First, these methods do not know \emph{when} to optimize.
During denoising, objects and their spatial layout are largely established in the early stages, while the later steps mainly refine low-level visual details~\citep{choi2022perception,hertz2023prompt}.
Existing methods adapt optimization across denoising steps using predefined schedules~\citep{li2025mixgrpo}, timestep-dependent rewards~\citep{savani2026stepwise,deng2026densegrpo,tong2026alleviating}, or attention-entropy signals~\citep{li2026aegpo}, but none of them locates the step at which a given model has fixed its layout, although this step differs across models (\Cref{sec:diag-crystal}).
We therefore measure, for each model, this transition from layout formation to detail refinement, and use it to set the SDE window where RL exploration takes place.

Second, current methods rely predominantly on preference models such as PickScore \citep{kirstain2023pick} for reward estimation.
On the latest SOTA open-source DMs \citep{wu2025qwenimage, flux-2-2025}, we find that preference scores are extremely high and nearly indistinguishable, providing negligible advantage for RL.
This calls for \textbf{structured, verifiable, and discriminative rewards}.

\begin{figure}[t]
    \centering
    \includegraphics[width=\textwidth]{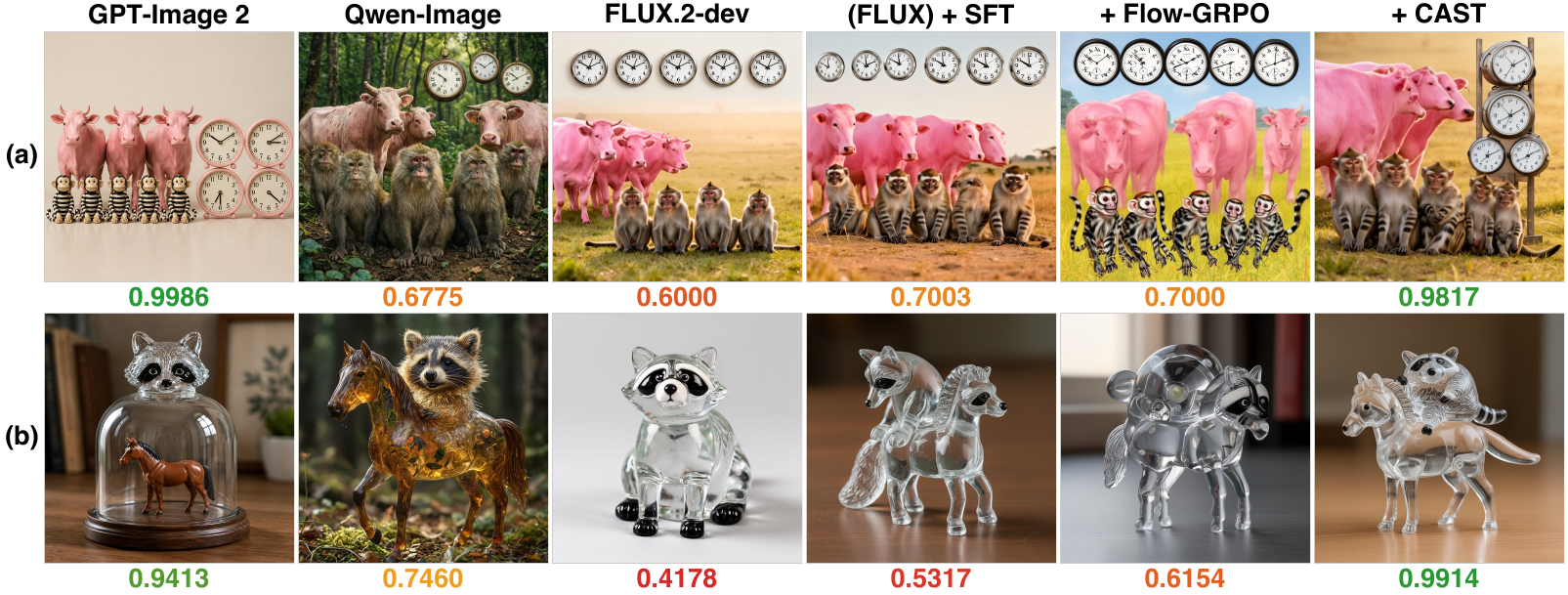}
    \caption{\textbf{Compositional failures in baselines and CAST's improvement.}
    GenEval~2 prompts:
    (a)~``five striped monkeys in front of three pink cows to the left of four clocks.''
    (b)~``a plastic horse under a glass raccoon.''
    GenEval~2 score reported below each image.
    Only CAST satisfies all verifiable-atoms in both prompts.
    More comparisons are provided in Appendix~\ref{app:qualitative}.}
    \label{fig:teaser}
\end{figure}

Third, a scalar reward collapses different failure modes into almost identical scores, providing little gradient signal for targeted improvement.
On GenEval~2 \citep{kamath2025geneval}, each prompt decomposes into 3--10 \textbf{\emph{verifiable-atoms}} (median 6.5), \ie, minimal, independently verifiable semantic units of a prompt covering object presence, count, attribute, action, and spatial relation.
Following the benchmark, we use the term \emph{\textbf{verifiable-atom}} throughout this paper.
A scalar reward, from a preference model or a sum of per-atom rewards, barely reveals \emph{\textbf{which}} verifiable-atom failed.

These three observations call for two measurements, which our diagnostic experiments (\Cref{sec:diagnostic}) provide.
To address the first limitation, we use a Tweedie estimate~\citep{efron2011tweedie} of the clean image at every step to track when objects and their spatial layout become established in each model.
To address the second and third, we score each verifiable-atom separately and localize it to specific image pixels via a teacher-forced attention heatmap~\citep{chefer2021generic} within a frozen vision-language model (VLM), Qwen3-VL~\citep{qwen2025qwen3vl}, yielding a \textbf{pixel-level granularity reward}.

Based on this analysis, we propose \textbf{CAST} (\textbf{C}ausal \textbf{A}dvantage-\textbf{S}tructured \textbf{T}raining), an RL fine-tuning method for pretrained DMs consisting of three stages (\Cref{fig:pipeline}).
(1)~\emph{CSG probe construction:} training prompts are synthesized from Causal Scene Graphs (CSG) \citep{yu2025lina}, where every node and edge is paired with a verifiable-atom probe (\Cref{subsec:method_probe}).
(2)~\emph{Atom-level scoring and grounding:} for each group of images sampled within the measured SDE window, the frozen VLM scores each atom independently and locates its evidence through teacher-forced attention (\Cref{subsec:method_scoring}).
(3)~\emph{Spatially weighted optimization:} CAST normalizes each atom's score within the group into an advantage, projects these advantages into a per-pixel advantage map, and uses the map to weight the SDE policy objective (\Cref{subsec:method_objective}).
As illustrated in \Cref{fig:teaser}, CAST produces images that satisfy all verifiable-atoms where existing methods fail.

We apply CAST to FLUX.2-dev and Qwen-Image-2512, two of the strongest open-source DMs, whose preference scores are already near saturation (\Cref{sec:diag-saturation}); improving them is therefore much harder than improving the weaker models used in earlier diffusion RL work.
We evaluate on GenEval~2, which checks every verifiable-atom of compositionally complex prompts, and on Qwen-Image-Bench~\citep{li2026qib}, which measures overall generation quality and thus reveals whether compositional gains come at the cost of image quality.
The base models already score $\geq 0.95$ on $11.6\%$--$18.8\%$ of GenEval~2 prompts, leaving no room for improvement there and diluting the full-set average; we therefore focus on a Hard Case subset of prompts on which the base models still fail.
On this subset, with the same budget and reward signal, CAST improves over the base model by $1.93\times$--$3.07\times$ as much as Flow-GRPO with a scalar reward, while achieving the best overall quality on Qwen-Image-Bench.

Our contributions are:
\begin{itemize}[leftmargin=2em]
    \item We show that preference rewards saturate on strong DMs and that a scalar reward conflates distinct failures; modeling prompts with CSG as separately scored verifiable-atoms restores a discriminative signal that tells the model \emph{which} part failed.
    \item We propose CAST, which applies independently normalized verifiable-atom advantages within a per-model SDE window, so that optimization targets the steps where layout is decided.
    \item We introduce the per-pixel advantage map, which projects atom-level advantages into pixel space via VLM attention heatmaps, so that gradients reach the image regions responsible for each success or failure instead of the whole image.
\end{itemize}

\section{Preliminaries}\label{sec:preliminaries}

\subsection{Flow Matching and Marginal-Preserving SDE Sampling}

Following the rectified-flow convention \citep{liu2023rectifiedflow, esser2024scaling}, we sample $(\vx_0,\vc) \sim p_{\mathrm{data}}$, $\vx_1 \sim \mathcal{N}(\mathbf{0},\mI)$, and $t \sim \mathcal{U}[0,1]$, with $\vx_t=(1-t)\vx_0+t\vx_1$. The velocity network defines the reference flow through
\begin{equation}
    \mathcal{L}_{\mathrm{FM}}(\theta)=\mathbb{E}\!\left[\left\|\vv_\theta(\vx_t,t,\vc)-(\vx_1-\vx_0)\right\|_2^2\right],
    \qquad \frac{\mathrm{d}\vx_t}{\mathrm{d}t}=\vv_\theta(\vx_t,t,\vc).
\end{equation}
The ODE is integrated from $t{=}1$ to $t{=}0$. For online RL, Flow-GRPO replaces this deterministic rollout with a reverse-time SDE that preserves its continuous-time marginals \citep{liu2025flowgrpo}:
\begin{equation}
    \mathrm{d}\vx_t=\left[\vv_t(\vx_t,\vc)-\frac{\sigma_t^2}{2}\nabla_{\vx}\log p_t(\vx_t\mid\vc)\right]\mathrm{d}t+\sigma_t\,\mathrm{d}\vw_t.
\end{equation}
Here $\sigma_t$ controls the noise level. For the rectified-flow path, $\nabla_{\vx}\log p_t(\vx_t\mid\vc)=-[\vx_t+(1-t)\vv_t(\vx_t,\vc)]/t$, so no separate score model is needed. Euler--Maruyama gives the Gaussian reverse transition
\begin{equation}
    \pi_\theta(\vx_{t-\Delta t}\mid\vx_t,\vc)=\mathcal{N}\!\left(\vx_{t-\Delta t};\mu_\theta(\vx_t,t,\vc),\,\sigma_t^2\Delta t\,\mI\right),
\end{equation}
where $\mu_\theta$ is the Euler mean. This density enables transition ratios and KL regularization; CAST uses it inside its calibrated window and follows the ODE outside.

\subsection{Group Relative Policy Optimization (GRPO)}

GRPO \citep{shao2024deepseekmath, liu2025flowgrpo} avoids a value model by normalizing rewards within a group. For $G$ outputs under the same $\vc$, let $R_i=R(\vx_0^{(i)},\vc)$, $\mu_G=G^{-1}\sum_{i=1}^{G}R_i$, $\sigma_G=[G^{-1}\sum_{i=1}^{G}(R_i-\mu_G)^2]^{1/2}$, and $\hat{A}^{(i)}=(R_i-\mu_G)/(\sigma_G+\epsilon)$. These advantages enter a clipped policy-gradient surrogate through the Gaussian transition ratios above; CAST replaces the image-level advantage with a per-atom, spatially weighted one.

\section{Diagnostic of Three Limitations in Current DM RL}\label{sec:diagnostic}

\begin{figure}[t]
    \centering
    \includegraphics[width=\textwidth]{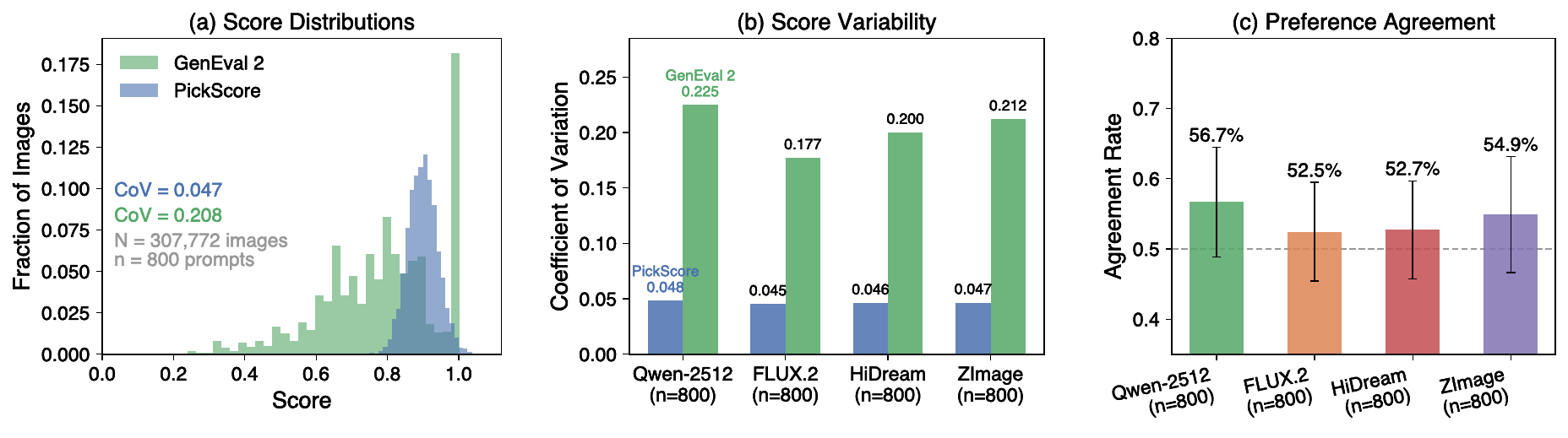}
    \caption{\textbf{PickScore saturates and weakly tracks structured correctness.}
    (a) Score distributions pooled across the four DMs: PickScore concentrates near its ceiling, whereas GenEval~2 retains a broad dynamic range.
    (b) The contrast persists within every model's generations.
    (c) Preference agreement between PickScore and GenEval~2 is only slightly above the 50\% random baseline (dashed line); error bars show prompt-level standard deviations.}
    \label{fig:metric}
\end{figure}

We identify three limitations of the current diffusion RL paradigm that collectively motivate CAST's design.
Each is supported by empirical evidence on state-of-the-art open-source DMs.

\subsection{Image Structure Is Established Early During Denoising}\label{sec:diag-crystal}

To locate when the semantic structure of the output image is determined, we compute Tweedie estimates~\citep{efron2011tweedie} $\hat{\vx}_0^{(t)}$ (one-step predictions of the clean image) at every denoising step and evaluate CLIP-ViT-L/14 \citep{radford2021learning} similarity between the estimate and the prompt, with $\tau\in[0,1]$ denoting progress from the initial noise to the final image.

CLIP similarity rises steeply during the first 10--20\% of denoising and then plateaus.
The point of maximum curvature (visualizations and detection details in Appendix~\ref{app:knee}, Figure~\ref{fig:crystal}) occurs at $\tau \approx 0.11$ for SD3.5-Large, $\tau \approx 0.21$ for FLUX.2-dev, and $\tau \approx 0.14$ for Qwen-Image-2512, where the decoded estimates already show the main layout and spatial composition.
Late SDE perturbations therefore risk disrupting established content, while the measured transition provides a model-specific signal for setting the SDE window rather than treating all timesteps uniformly.

\subsection{Preference Models Saturate on Strong DMs}\label{sec:diag-saturation}

Current diffusion RL methods rely predominantly on \textbf{preference models} such as PickScore \citep{kirstain2023pick} and HPSv2 \citep{wu2023human}.
We compare PickScore against GenEval~2 \citep{kamath2025geneval}, which verifies compositional correctness atom by atom, on a shared corpus of 307,772 images generated by four strong DMs: Qwen-Image-2512~\citep{wu2025qwenimage}, FLUX.2-dev~\citep{flux-2-2025}, HiDream~\citep{cai2026hidream}, and ZImage~\citep{team2025zimage}.

As shown in Figure~\ref{fig:metric}(a--b), PickScore is narrowly concentrated: its per-model mean ranges from 0.887 to 0.907 \bad{(high saturation)}, with a coefficient of variation (CoV) of only 0.045--0.048 \bad{(low discriminability)}.
In contrast, GenEval~2 produces per-model means of 0.734--0.829 \good{(low saturation)} and CoVs of 0.177--0.225 \good{(high discriminability)}, preserving substantially greater score variation for advantage estimation.
Preference models are thus ineffective for advantage estimation on strong DMs, calling for structured evaluation of atom failures.

\subsection{Scalar Reward Is Blind to Structural Correctness}\label{sec:diag-blind}

Even when scoring systems produce non-saturated scores, a holistic score does not separately check each verifiable-atom specified by the prompt.
On the same corpus, we pair the images generated under the same prompt and measure whether PickScore and GenEval~2 rank the two images of each pair in the same order (Figure~\ref{fig:metric}(c)).
Across the four models, agreement stays between 52.5\% and 56.7\%, only slightly above the 50\% random baseline.
This weak correspondence indicates that holistic preference and structured compositional correctness rank generated images differently.

This result shows that a scalar reward conflates distinct failure modes into nearly identical values.
An image with correct objects but incorrect spatial relations and another with correct relations but missing objects may receive the same reward, obscuring which error should be corrected.
This limitation is not unique to preference models: although GenEval~2 exposes separate scores for different verifiable-atoms, collapsing them into a single aggregate reward would recreate the same credit-assignment ambiguity.
The key issue is therefore not only which evaluator is used, but whether its structured outputs are preserved throughout optimization.
Together, these findings motivate three design choices: (1) use the measured transition to set the SDE window, (2) retain independently verifiable atom-level reward factors, and (3) spatially ground each factor in the image regions containing its evidence. We implement all three in \textbf{CAST}.

\section{Method}\label{sec:method}

\begin{figure}[t]
    \centering
    \includegraphics[width=\textwidth]{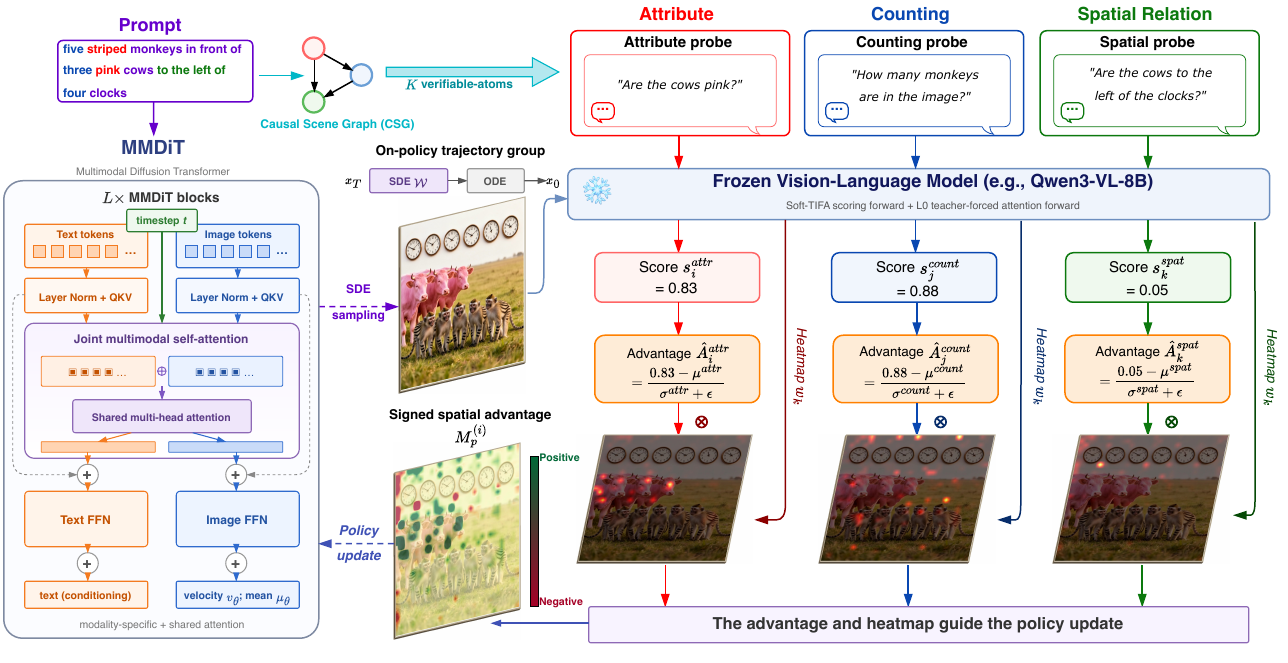}
    \caption{\textbf{Overview of CAST.}
    A Causal Scene Graph (CSG) decomposes the prompt into $K$ verifiable-atoms.
    Three representative types of facts are shown: \textcolor{red}{attributes} (red), \textcolor{blue}{counts} (blue), and \textcolor{teal}{spatial relations} (green).
    A frozen VLM uses separate passes to score each atom and extract its teacher-forced attention map.
    The resulting atom advantages and maps are combined into a signed spatial map that weights the SDE policy objective.}
    \label{fig:pipeline}
\end{figure}

In \textbf{CAST}, different verifiable-atoms independently direct training gradients to the image regions where their evidence resides, through three stages (Figure~\ref{fig:pipeline}): CSG probe construction (\Cref{subsec:method_probe}), per-atom scoring and grounding (\Cref{subsec:method_scoring}), and spatially weighted optimization (\Cref{subsec:method_objective}).

\subsection{VQA Probe Construction via Causal Scene Graphs}\label{subsec:method_probe}

The Causal Scene Graph (CSG) \citep{yu2025lina} decomposes a text prompt into a directed graph $G = (V, E)$, where $V$ represents semantic entities and $E$ encodes relations among entity pairs.
Each node, node attribute, and edge maps to one VQA probe question: (1) node existence $\to$ object questions; (2) node attributes $\to$ property questions; (3) edges $\to$ relational questions.
For example, the prompt ``four white bicycles in front of three plastic cows'' yields nodes \{bicycle, cow\} with attributes [white, count=4] and [plastic, count=3], and a spatial edge (bicycle $\xrightarrow{\text{in front of}}$ cow).

In CAST, the CSG serves as our data construction method rather than an analysis tool: we sample a graph first and render it into a prompt.
By controlling the variables in the graph, such as swapping entities, attributes, counts, and spatial relations, we can scale the dataset while keeping a clean causal structure among its verifiable-atoms.
Each of the $K$ atoms is paired with a VQA probe and the words in the prompt that states the corresponding fact before any image is generated.

\subsection{Per-Atom Scoring and Spatial Grounding}\label{subsec:method_scoring}

\paragraph{Soft scoring.}
Each verifiable-atom is scored using the GenEval~2 Soft-TIFA protocol \citep{kamath2025geneval}, with Qwen3-VL-8B \citep{qwen2025qwen3vl} as the frozen scorer.
The VLM receives the image and question, and the score sums the softmax probabilities of all answer variants:
\begin{equation}\label{eq:soft-tifa}
    s_k = \sum_{v \in \mathcal{V}_k} P(v \mid x, q_k),
\end{equation}
where $\mathcal{V}_k$ contains the accepted token variants of the correct answer (e.g., capitalization and tokenization variants of ``yes'' for binary questions, and word and digit forms for count questions).

\paragraph{Spatial grounding via teacher-forced attention.}
While $s_k$ tells \emph{whether} atom $k$ is correct, it provides no information about \emph{where} in the image the evidence resides.
To obtain spatial grounding, we run a separate teacher-forced decoding pass.
The VLM receives the image, while the original prompt is supplied as the target response.
The CSG maps each atom to the words in the prompt that states the corresponding fact.
At the first decoder layer (L0), we average the attention heads and the selected prompt-token query rows, retain their attention to visual tokens, and normalize it to unit spatial mean at pixel space.
This yields a heatmap $w_k$ for every atom in one forward pass.

\subsection{Spatially Weighted SDE Policy Optimization}\label{subsec:method_objective}

\paragraph{Windowed SDE sampling.}
Given a prompt $\vc$, we sample a group of $N$ reverse-time trajectories from the current policy.
The measured structure-formation point defines a model-specific early candidate range, from which each trajectory uniformly samples a contiguous two-transition SDE window; the exact discretization is provided in Appendix~\ref{app:knee}.
Let $\Delta t>0$ denote the magnitude of one reverse-time step.
Within it, we use the Flow-GRPO Gaussian transition:
\begin{equation}
    \pi_\theta(\vx_{t-\Delta t}\mid \vx_t,\vc)
    =
    \mathcal{N}\!\left(
    \vx_{t-\Delta t};
    \mu_\theta(\vx_t,t,\vc),
    \sigma_t^2\Delta t\,\mI
    \right),
    \qquad t\in\mathcal{W}.
\end{equation}
Outside $\mathcal{W}$, sampling follows the original deterministic ODE.
We retain the transition probabilities inside $\mathcal{W}$ for policy optimization and evaluate the resulting final images with the per-atom scorer.

\paragraph{Z-score advantage.}
For $N$ images of one prompt, each atom $k$ has advantage:
\begin{equation}\label{eq:cast_advantage}
    \hat{A}_k^{(i)} = \operatorname{clip}\!\left(
    \frac{s_k^{(i)} - \mu^{(k)}}{\sigma^{(k)} + \epsilon},
    -5,\,5
    \right),
\end{equation}
where $\mu^{(k)}$ and $\sigma^{(k)}$ are the mean and standard deviation of $s_k$ across the group.
If the score variation is below the scorer's numerical resolution, we set the corresponding advantages to zero.

\paragraph{Per-pixel advantage map.}
Each atom's scalar advantage $\hat{A}_k$ and spatial heatmap $w_k$ combine to project the advantage into pixel space.
Each VLM heatmap is first resized to the generated image.
For policy optimization, we average the heatmap over the pixels corresponding to each denoiser patch $p$, yielding $w_{k,p}$.
The combined advantage map is
\begin{equation}\label{eq:pixel_advantage}
    M_p^{(i)}
    = \operatorname{clip}\!\left(
    \sum_{k=1}^{K} \hat{A}_k^{(i)} w_{k,p}^{(i)},
    -5,\,5
    \right).
\end{equation}
The signed contributions are summed before clipping, allowing positive and negative atom advantages to offset each other at the same spatial location.

\paragraph{Spatial transition ratios.}
The diagonal Gaussian log-density separates over latent dimensions.
We therefore retain the patch dimension and average only over the $D$ channels within each patch:
\begin{equation}\label{eq:spatial_ratio}
\begin{aligned}
    \ell_{\theta,t,p}^{(i)}
    &= \frac{1}{D}\sum_{d=1}^{D}
    \log \mathcal{N}\!\left(
    x_{t-\Delta t,p,d}^{(i)};
    \mu_{\theta,t,p,d}^{(i)},
    \sigma_t^2\Delta t
    \right), \\
    r_{t,p}^{(i)}(\theta)
    &= \exp\!\left(
    \ell_{\theta,t,p}^{(i)}
    - \ell_{\mathrm{old},t,p}^{(i)}
    \right).
\end{aligned}
\end{equation}

CAST uses the resulting normalized patch ratios in the following clipped spatial policy objective:
\begin{equation}\label{eq:cast_loss}
\begin{aligned}
    \mathcal{L}_{\mathrm{CAST}}
    ={}& -\mathbb{E}_{i,\,t\in\mathcal{W},\,p}
    \Big[\min\big(
    r_{t,p}^{(i)} M_p^{(i)},
    \operatorname{clip}(r_{t,p}^{(i)},1-\varepsilon_{\mathrm{clip}},1+\varepsilon_{\mathrm{clip}})M_p^{(i)}
    \big)\Big] \\
    &+ \beta\,\mathbb{E}_{i,\,t\in\mathcal{W}}
    \left[D_{\mathrm{KL}}(\pi_\theta\,\|\,\pi_{\mathrm{ref}})\right].
\end{aligned}
\end{equation}
The reference policy disables the LoRA adapters; because both Gaussian transitions share the same covariance, the KL term has the closed-form expression of Flow-GRPO \citep{liu2025flowgrpo}.

Positive values of $M_p$ increase the likelihood of the sampled local transition, while negative values decrease it.
The factorization applies to the conditional Gaussian density, not to the denoiser itself: every local mean remains a function of the complete state $\vx_t$ through the globally coupled MMDiT.

\section{Experiments}\label{sec:experiments}

\subsection{Experimental Setup}\label{sec:baselines}

All trainable methods share a single pool of 1,024 training prompts synthesized by sampling Causal Scene Graphs (\Cref{subsec:method_probe}): entities endowed with counts and attributes are composed through spatial relations, and the sampled graph is rendered into fluent text.
Every node, attribute, and edge is emitted together with its verifiable-atom probe and the words in the prompt, so all $K$ atoms are fixed by construction before any image is generated.
The pool is disjoint from both evaluation benchmarks in exact and normalized form, and all checkpoint and hyperparameter selection uses a held-out 256-prompt development split; the benchmarks are never used for selection.

\paragraph{Baselines.}
We evaluate CAST on FLUX.2-dev~\citep{flux-2-2025} and Qwen-Image-2512~\citep{wu2025qwenimage} against Base, two SFT variants, and two Flow-GRPO variants.
\textbf{Base} is the unmodified pretrained model.
\textbf{SFT (4/8)} and \textbf{SFT (3/64)} are self-distillation baselines: for each pool prompt, the backbone generates 8 (resp.\ 64) candidates, the frozen verifier scores them under the GenEval~2 protocol, and the four (resp.\ three) highest-scoring candidates are retained and fine-tuned with uniform weights for three epochs.
SFT (3/64) serves as a high-selectivity distillation upper bound.
\textbf{Flow-GRPO}~\citep{liu2025flowgrpo} is evaluated with its original aesthetic reward and an atom-aggregated reward under the GenEval~2 (G2) protocol; both train on-policy on the same pool.

\paragraph{Experimental setting.}
Within each backbone, all trainable methods share the LoRA \citep{hu2022lora} configuration and initialization.
The two Flow-GRPO variants and CAST further share the same 128-prompt manifest, prompt order, training seed, SDE window, and learning-rate schedule; each performs 128 single-prompt collections with group size 16, matching the 2,048 generated images and 128 optimizer updates of Table~\ref{tab:main_results}(b).
Implementation and inference details are provided in Appendix~\ref{app:implementation_details}.

\subsection{Evaluation}\label{sec:evaluation}

We use \textbf{GenEval~2}~\citep{kamath2025geneval} to measure compositional prompt adherence and \textbf{Qwen-Image-Bench}~\citep{li2026qib} to measure overall generation quality.
We evaluate on all 800 GenEval~2 prompts and all 1,000 Qwen-Image-Bench prompts, sharing with the training pipeline only the scoring protocol and the frozen verifier (\Cref{sec:baselines}).

\paragraph{Hard Case subset.}
Although GenEval~2 is far more discriminative than preference rewards in aggregate (\Cref{sec:diag-saturation}), a nontrivial fraction of its prompts is already saturated even for the untrained base models: averaged over the four evaluation seeds, the per-prompt Overall score is $\geq 0.95$ on 150 of 800 prompts (18.8\%) for FLUX.2-dev and 93 of 800 (11.6\%) for Qwen-Image-2512.
Near-ceiling prompts contribute almost identical scores to every method, so the full-set average understates the remaining differences.
We therefore report a 50-prompt \textbf{Hard Case} subset alongside the full-set average: the frozen FLUX.2-dev base model generates 64 images per evaluation prompt, the frozen verifier scores them for subset selection only, and we retain the 100 highest-variance prompts before keeping the 50 with the lowest mean score (\Cref{app:evaluation_protocol}).

All methods use the same inference settings and matched prompt-specific seeds.
Each frozen checkpoint is evaluated with four matched generation-seed replicates, and we report the mean and sample standard deviation across replicates.
The complete protocol is provided in Appendix~\ref{app:evaluation_protocol}.

\subsection{Main Results}\label{sec:main_results}

\begin{table*}[t]
    \centering
    \caption{\textbf{Main performance and training cost.}
    \textbf{(a)} Final held-out performance on GenEval~2 and Qwen-Image-Bench.
    G2 Hard-50 is the GenEval~2 Overall score on the 50-prompt Hard Case subset of unsaturated evaluation prompts (\Cref{sec:evaluation}).
    Values are mean $\pm$ sample standard deviation over four matched generation-seed replicates.
    \textbf{(b)} Raw training cost within each backbone. All generated and scored candidates, including filtered or discarded candidates, are counted.}
    \label{tab:main_results}
    \small
    \setlength{\tabcolsep}{2.2pt}

    \textbf{(a) Final performance}\\[2pt]
    \begin{tabular}{@{}llccccc@{}}
        \toprule
        Backbone & Method & \makecell[tc]{G2\\Hard-50} $\uparrow$ & \makecell[tc]{G2\\Overall} $\uparrow$ & \makecell[tc]{G2\\Count} $\uparrow$ & \makecell[tc]{G2\\Position} $\uparrow$ & \makecell[tc]{QIB\\Overall} $\uparrow$ \\
        \midrule
        \multirow{6}{*}{\shortstack[l]{FLUX.2-\\dev}} & Base & $65.72 \pm 1.83$ & $83.16 \pm 0.20$ & $66.15 \pm 0.60$ & $73.29 \pm 0.42$ & $52.84 \pm 0.27$ \\
        & SFT (4/8) & $67.28 \pm 2.06$ & $83.31 \pm 0.24$ & $66.37 \pm 0.60$ & $73.46 \pm 0.51$ & $52.76 \pm 0.25$ \\
        & SFT (3/64) & $69.76 \pm 2.48$ & $83.93 \pm 0.40$ & $67.27 \pm 0.60$ & $74.15 \pm 0.88$ & $52.42 \pm 0.16$ \\
        & {\scriptsize Flow-GRPO (Orig.)} & $64.26 \pm 1.76$ & $81.85 \pm 0.22$ & $64.18 \pm 0.44$ & $72.71 \pm 1.13$ & $53.34 \pm 0.28$ \\
        & {\scriptsize Flow-GRPO (G2)} & $71.37 \pm 1.90$ & $84.54 \pm 0.11$ & $68.39 \pm 0.32$ & $74.98 \pm 0.62$ & $52.73 \pm 0.06$ \\
        & CAST & $\mathbf{76.64 \pm 2.39}$ & $\mathbf{85.95 \pm 0.10}$ & $\mathbf{70.85 \pm 0.29}$ & $\mathbf{76.90 \pm 0.53}$ & $\mathbf{53.44 \pm 0.29}$ \\
        \midrule
        \multirow{6}{*}{\shortstack[l]{Qwen-\\Image-\\2512}} & Base & $67.79 \pm 1.81$ & $78.02 \pm 0.56$ & $67.14 \pm 1.59$ & $59.60 \pm 0.84$ & $50.68 \pm 0.23$ \\
        & SFT (4/8) & $68.55 \pm 1.79$ & $78.27 \pm 0.54$ & $67.64 \pm 1.55$ & $59.62 \pm 0.78$ & $50.45 \pm 0.24$ \\
        & SFT (3/64) & $70.12 \pm 1.84$ & $79.29 \pm 0.45$ & $69.64 \pm 1.37$ & $59.72 \pm 0.53$ & $49.55 \pm 0.29$ \\
        & {\scriptsize Flow-GRPO (Orig.)} & $68.60 \pm 1.75$ & $77.90 \pm 0.44$ & $67.15 \pm 1.07$ & $59.92 \pm 0.91$ & $50.11 \pm 0.26$ \\
        & {\scriptsize Flow-GRPO (G2)} & $69.11 \pm 1.56$ & $79.59 \pm 0.22$ & $70.48 \pm 0.62$ & $58.59 \pm 0.25$ & $50.70 \pm 0.15$ \\
        & CAST & $\mathbf{71.84 \pm 1.85}$ & $\mathbf{80.81 \pm 0.23}$ & $\mathbf{72.35 \pm 0.55}$ & $\mathbf{60.45 \pm 0.43}$ & $\mathbf{51.32 \pm 0.15}$ \\
        \bottomrule
    \end{tabular}

    \vspace{0.7em}
    \textbf{(b) Training cost}\\[2pt]
    \begin{tabular}{@{}lrrrr@{}}
        \toprule
        Method & Images $\downarrow$ & Updates $\downarrow$ & FLUX H200-h $\downarrow$ & Qwen H200-h $\downarrow$ \\
        \midrule
        SFT (4/8) & 8,192 & 384 & 95.00 & 90.24 \\
        SFT (3/64) & 65,536 & 288 & 749.51 & 723.70 \\
        Flow-GRPO (Orig.) & 2,048 & 128 & 28.91 & 28.75 \\
        Flow-GRPO (G2) & 2,048 & 128 & 32.43 & 32.21 \\
        CAST & 2,048 & 128 & 32.90 & 32.51 \\
        \bottomrule
    \end{tabular}
\end{table*}

\paragraph{Main comparison.}
Flow-GRPO with its original preference reward provides no consistent improvement over the base models, while selective-distillation SFT yields only modest gains despite using up to $32\times$ more generated images.
Replacing the aesthetic reward with the aggregated GenEval~2 reward makes Flow-GRPO improve on both backbones and provides the matched-reward reference for CAST.
These full-set averages, however, compress the differences on the $11.6\%$--$18.8\%$ of evaluation prompts that are already saturated for the base models (\Cref{sec:evaluation}); we therefore focus on the Hard Case subset, with full-set scores reported in Table~\ref{tab:main_results}.
Within the same 2,048-image and 128-update budget and under the same reward signal, the Hard Case gain of CAST over the base model is $1.93\times$ that of scalar-reward Flow-GRPO on FLUX.2-dev and $3.07\times$ on Qwen-Image-2512.
CAST also surpasses the high-selectivity distillation upper bound SFT (3/64) on both backbones, whose Hard Case gain amounts to only $37\%$ and $58\%$ of CAST's despite its $32\times$ larger image budget, while attaining the highest Qwen-Image-Bench Overall on both backbones.

\paragraph{Ablations.}\label{sec:ablation}

\begin{figure}[t]
    \centering
    \includegraphics[width=\textwidth]{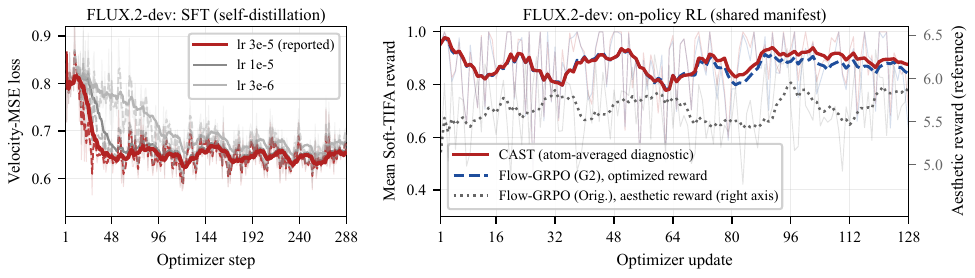}
    \caption{\textbf{Training dynamics on FLUX.2-dev.}
    \emph{Left:} SFT self-distillation loss under three learning rates; the reported setting is emphasized, and all runs plateau within the budget.
    \emph{Right:} mean Soft-TIFA reward on the shared RL manifest; the CAST curve is an atom-averaged diagnostic rather than its optimized objective, and the aesthetic reward (right axis) is on a separate scale.}
    \label{fig:training}
\end{figure}

Table~\ref{tab:ablation} isolates atom-level credit and spatial routing under the same on-policy training setting as CAST.
All variants use the same SDE window, training budget, and optimization settings.
Figure~\ref{fig:training} additionally shows the training dynamics of the SFT and RL runs on FLUX.2-dev.

\begin{table*}[t]
    \centering
    \caption{\textbf{Ablation of atom-level credit assignment and spatial weighting on FLUX.2-dev.}
    All variants use the same training prompts, SDE window, generated-image budget, and optimization settings. Values are mean $\pm$ sample standard deviation over four matched generation-seed replicates.}
    \label{tab:ablation}
    \small
    \setlength{\tabcolsep}{5pt}
    \begin{tabular}{@{}lllcc@{}}
        \toprule
        Variant & Advantage & Spatial map & G2 Overall $\uparrow$ & QIB Overall $\uparrow$ \\
        \midrule
        Sum-reward baseline & per-image & none & $84.54 \pm 0.11$ & $52.73 \pm 0.06$ \\
        Per-atom advantage & per-atom & none & $85.12 \pm 0.14$ & $52.98 \pm 0.18$ \\
        Spatial weighting & per-image & aggregated attention & $84.91 \pm 0.16$ & $53.06 \pm 0.20$ \\
        Shuffled correspondence & per-atom & shuffled atom maps & $85.00 \pm 0.15$ & $52.89 \pm 0.22$ \\
        \textbf{CAST} & \textbf{per-atom} & \textbf{matched atom maps} & $\mathbf{85.95 \pm 0.10}$ & $\mathbf{53.44 \pm 0.29}$ \\
        \bottomrule
    \end{tabular}
\end{table*}

\section{Related Work}

\paragraph{RL for Diffusion and Flow Matching.}
DDPO~\citep{black2024training} formulated diffusion denoising as a multi-step MDP with terminal rewards.
Flow-GRPO~\citep{liu2025flowgrpo} then extended GRPO~\citep{shao2024deepseekmath} to flow matching by converting the deterministic ODE into an equivalent SDE for stochastic exploration.
Later methods improved temporal credit assignment through per-step reward gains~\citep{savani2026stepwise} or adaptive timestep-wise stochasticity~\citep{deng2026densegrpo}.
Neighbor GRPO~\citep{he2025neighborgrpo} instead retained deterministic ODE sampling and perturbed only the initial noise.
CAST complements temporal credit assignment with spatial credit assignment: each atom advantage weights the corresponding evidence region within a calibrated early SDE window.

\paragraph{Compositional Generation and Evaluation.}
TIFA~\citep{hu2023tifa} evaluates prompt alignment using VQA questions generated from the prompt, while VQAScore~\citep{lin2024evaluating} uses a single global question.
GenEval~2~\citep{kamath2025geneval} provides a finer decomposition by evaluating prompt atoms across object, count, attribute, spatial relation, and verb skills.
It also introduces Soft-TIFA, which uses VLM token probabilities for continuous scoring.
LINA~\citep{yu2025lina} represents prompt content with causal scene graphs to diagnose physical alignment failures.
CAST uses it for training, projecting each atom's advantage onto its evidence.

\paragraph{Dense Reward for Generation.}
Preference models such as PickScore~\citep{kirstain2023pick}, ImageReward~\citep{xu2023imagereward}, and HPSv2~\citep{wu2023human} provide scalar rewards for diffusion-model training.
MPS~\citep{zhang2024learning} decomposes preference into several dimensions, but each remains a global score without spatial localization.
VisionReward~\citep{xu2026visionreward} uses fine-grained VQA questions, but likewise produces only a scalar score for each question.
Focus-N-Fix~\citep{xing2025focus} localizes problematic regions for region-aware fine-tuning, but uses a single global heatmap, whereas CAST pairs every atom advantage with its own map.

\section{Conclusion}\label{sec:conclusion}

We introduce \textbf{CAST}, which trains diffusion models with spatially grounded atom-level rewards.
Our diagnostics show that strong DMs fix their layout early, that preference rewards saturate on them, and that a scalar reward conflates distinct failures.
CAST therefore sets a model-specific early SDE window, computes a separate advantage for each verifiable-atom, and projects it via teacher-forced attention onto its image region to weight the policy objective.
On two strong DMs, CAST improves adherence on the hardest GenEval~2 prompts well beyond Flow-GRPO at equal budget while preserving overall quality, enabling finer-grained credit assignment than a scalar reward.

\clearpage

\bibliography{references}
\bibliographystyle{iclr2027_conference}

\clearpage

\appendix

\section{Additional Qualitative Comparisons}\label{app:qualitative}

\begin{figure}[t]
    \centering
    \includegraphics[width=\textwidth]{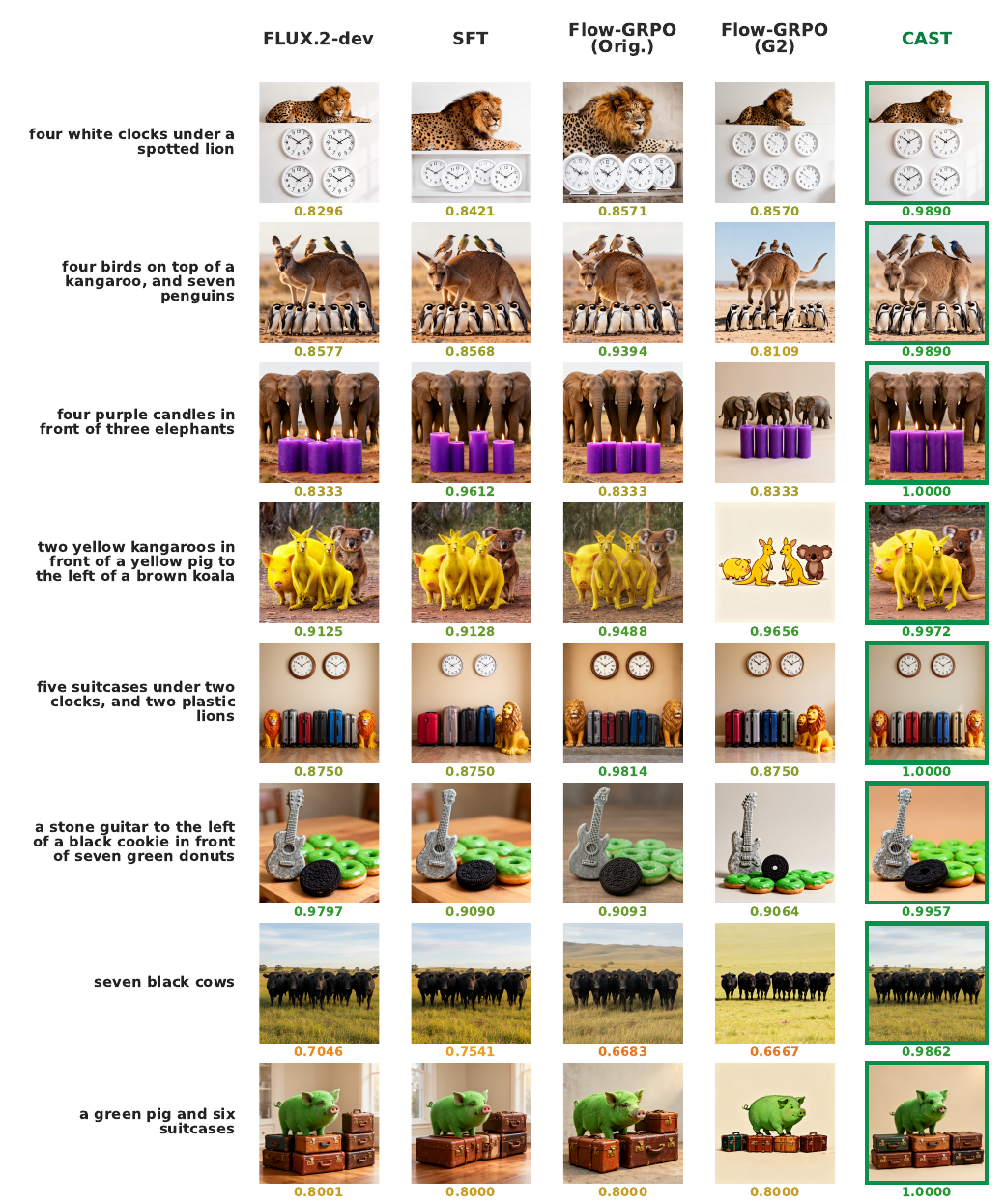}
    \caption{\textbf{Additional qualitative comparisons on FLUX.2-dev (part I).}
    GenEval~2 prompts with multi-object counts, attributes, and chained spatial relations.
    GenEval~2 score reported below each image.}
    \label{fig:qualitative-main}
\end{figure}

\begin{figure}[t]
    \centering
    \includegraphics[width=\textwidth]{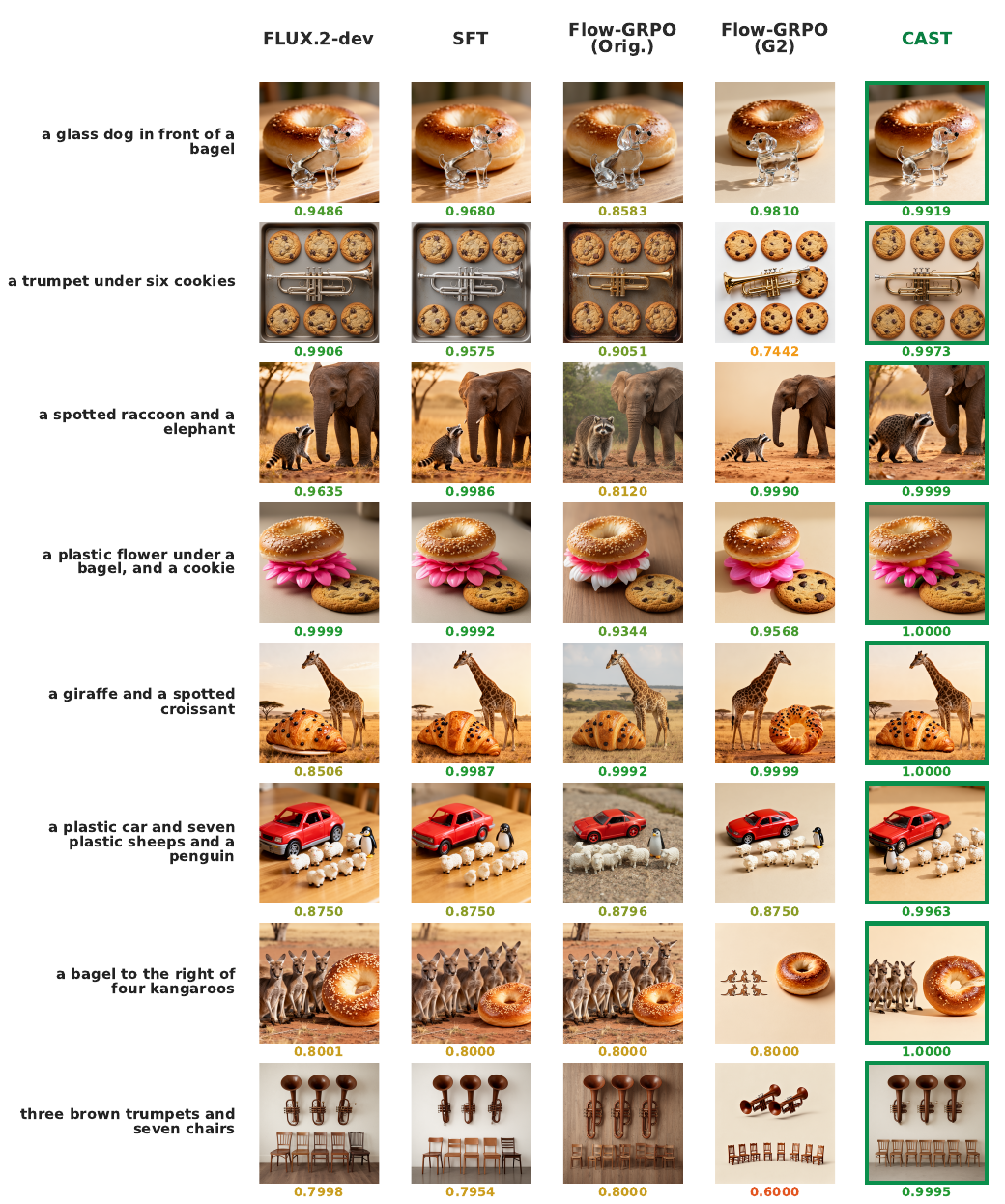}
    \caption{\textbf{Additional qualitative comparisons on FLUX.2-dev (part II).}
    GenEval~2 score reported below each image.}
    \label{fig:qualitative-backup}
\end{figure}

Figures~\ref{fig:qualitative-main}--\ref{fig:qualitative-backup} provide additional qualitative comparisons on FLUX.2-dev, complementing the teaser examples in \Cref{fig:teaser}.
Each row shows the images generated by the base model, the strongest SFT variant, both Flow-GRPO variants, and CAST for one GenEval~2 evaluation prompt, together with the per-prompt GenEval~2 score.

\section{Structure Formation Point Detection}\label{app:knee}

The CLIP similarity trajectory $s(\tau)$ exhibits a characteristic rise-then-plateau shape, where $\tau \in [0,1]$ denotes normalized denoising progress.
We define the structure formation point $\tau^*$ as the point of maximum curvature:
\begin{equation}
    \tau^* = \arg\max_\tau \; \kappa(\tau), \qquad
    \kappa(\tau) = \frac{|s''(\tau)|}{\bigl(1 + s'(\tau)^2\bigr)^{3/2}}\,.
\end{equation}
This is the knee of the curve: the point where the steep rise in semantic alignment bends into the flat refinement plateau.
In practice, we fit a degree-4 smoothing spline to the mean CLIP scores and evaluate $\kappa(\tau)$ on the resulting curve.
The three models yield $\tau^* \approx 0.11$ (SD3.5-Large), $\tau^* \approx 0.21$ (FLUX.2-dev), and $\tau^* \approx 0.14$ (Qwen-Image-2512), all before $\tau = 0.25$.

\begin{figure}[t]
    \centering
    \includegraphics[width=\textwidth]{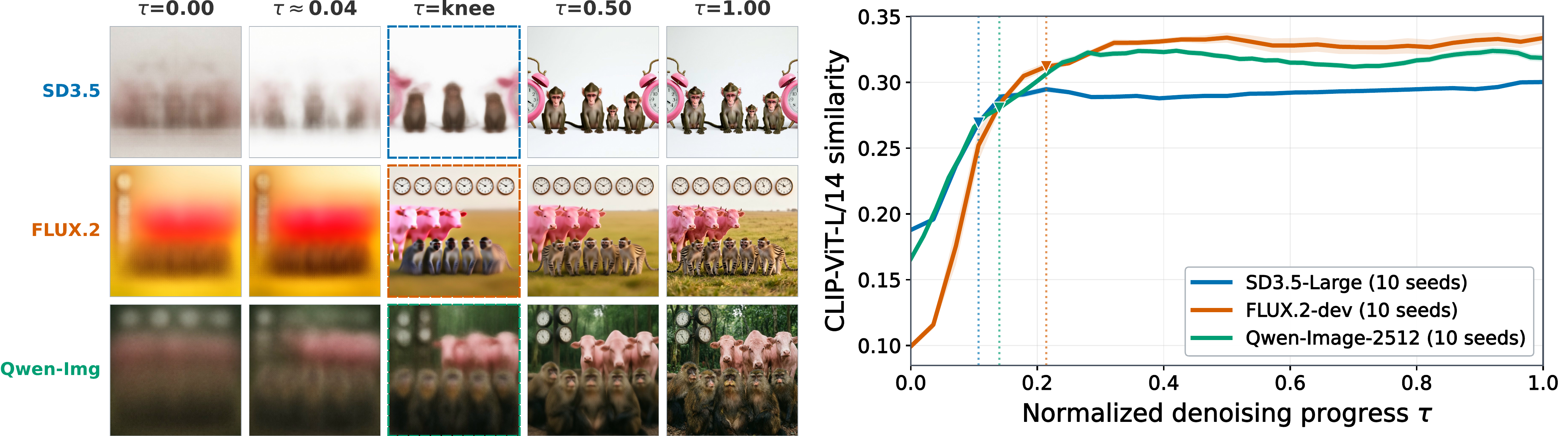}
    \caption{\textbf{Image structure forms in early denoising steps.}
    Left: intermediate decoded images at five timesteps for three diffusion models.
    Dashed borders mark the knee point where CLIP similarity plateaus.
    Right: CLIP-ViT-L/14 similarity between Tweedie estimates and the text prompt.
    The knee point (triangle) occurs before $\tau{=}0.25$ across all models, indicating that the main layout and composition are established early.}
    \label{fig:crystal}
\end{figure}

\paragraph{Discretization into the SDE window.}
For an $N$-step sampler, we set $b_m=\lceil\tau_m^*(N-1)\rceil$ and define the candidate transition range as $\mathcal{R}_m=[a_m,b_m)$.
We use $a_m=0$ for FLUX.2-dev and $a_m=1$ for Qwen-Image-2512, excluding Qwen's initial $\sigma=1$ transition to preserve numerical parity under differentiable FSDP recomputation.
With the fixed window length $L=2$, we sample
\begin{equation}
    s \sim \operatorname{Uniform}\{a_m,\ldots,b_m-L\},
    \qquad
    \mathcal{W}_s=[s,s+L).
\end{equation}
For $N=50$, this gives $\mathcal{R}_m=[0,11)$ for FLUX.2-dev and $\mathcal{R}_m=[1,7)$ for Qwen-Image-2512.
Only the two transitions in $\mathcal{W}_s$ are stochastic; all remaining transitions follow the deterministic ODE.

\section{Additional Ablation Details}\label{app:ablation}

The main ablation in Table~\ref{tab:ablation} uses the same on-policy setting as CAST.
The sum-reward baseline broadcasts one normalized per-image advantage uniformly, while the per-atom advantage variant normalizes each atom separately before broadcasting their sum.
Spatial weighting combines the per-image advantage with an aggregate attention map, and shuffled correspondence permutes the atom-to-map pairing.

\section{Implementation Details}\label{app:implementation_details}

\paragraph{Prompt pool construction.}
Each of the 1,024 training prompts is synthesized by sampling a Causal Scene Graph (\Cref{subsec:method_probe}): entities endowed with counts and attributes are composed through spatial relations, and the sampled graph is rendered into fluent text.
Every node, attribute, and edge is emitted together with its verifiable-atom probe and the words in the prompt, so all $K$ atoms are fixed by construction before any image is generated.
The pool mirrors the compositional structure that GenEval~2-style prompts instantiate, yet is generated from scratch: it is disjoint from the 800 GenEval~2 and 1,000 Qwen-Image-Bench evaluation prompts in both exact and normalized form.
All checkpoint and hyperparameter selection uses a held-out set of 256 development prompts from the same construction; the benchmarks are never used for selection.
Each RL method runs 128 single-prompt collections with group size 16, matching its 2,048 generated images and 128 optimizer updates (Table~\ref{tab:main_results}(b)).

\paragraph{Shared configuration.}
For each backbone, all trainable methods share the constructed prompt pool (\Cref{sec:baselines}), the LoRA target modules, rank, scaling factor, and initialization; the two Flow-GRPO variants and CAST further share the same 128-prompt training manifest, prompt order, training seed, SDE window, and learning-rate schedule.
Generated-image and optimizer-update budgets are reported separately for each method.

\paragraph{Flow-GRPO rewards.}
The Aesthetic variant uses the frozen LAION Improved Aesthetic Predictor, which applies the released MLP to CLIP ViT-L/14 image features.
For the G2 variant, the per-atom Soft-TIFA scores are averaged into one image-level reward before within-prompt group normalization and clipping.
It therefore uses neither per-atom advantages nor spatial maps.

\begin{table}[h]
\centering
\caption{\textbf{Settings shared by all compared methods.}}
\label{tab:implementation_details}
\begin{tabular}{ll}
\toprule
Setting & Value \\
\midrule
Image resolution & $1024\times1024$ \\
Sampling steps & 50 \\
Guidance & Backbone default \\
LoRA rank & 64 \\
LoRA scaling factor & $\alpha=64$ \\
Target modules & Matched within each backbone \\
Initialization & Matched within each backbone \\
\bottomrule
\end{tabular}
\end{table}

\section{Training Cost Accounting}\label{app:compute_accounting}

Generated/scored images count every candidate evaluated during training, including candidates later filtered or discarded.
Training updates count optimizer steps applied to the trainable denoiser.
H200-hours include image generation, reward scoring, attention extraction when required, and optimization.
Costs are compared only between methods using the same backbone.

\section{Attention-Head Diagnostic}\label{app:head_analysis}

We analyze the pre-projection attention-head activations of FLUX.2-dev on 10 hard GenEval~2 prompts with 64 generated images per prompt.
This diagnostic is not used by CAST during training.
We record activations from all 56 transformer blocks at denoising steps 3, 7, 14, and 21.
For each verifiable atom, we measure the normalized difference between the mean activations of correct and incorrect images, then aggregate the results by count, attribute, and spatial relation.
We estimate significance with 200 label permutations, combine evidence across prompts with Fisher's method, and apply Benjamini--Hochberg FDR correction.

\begin{figure}[t]
    \centering
    \includegraphics[width=\linewidth]{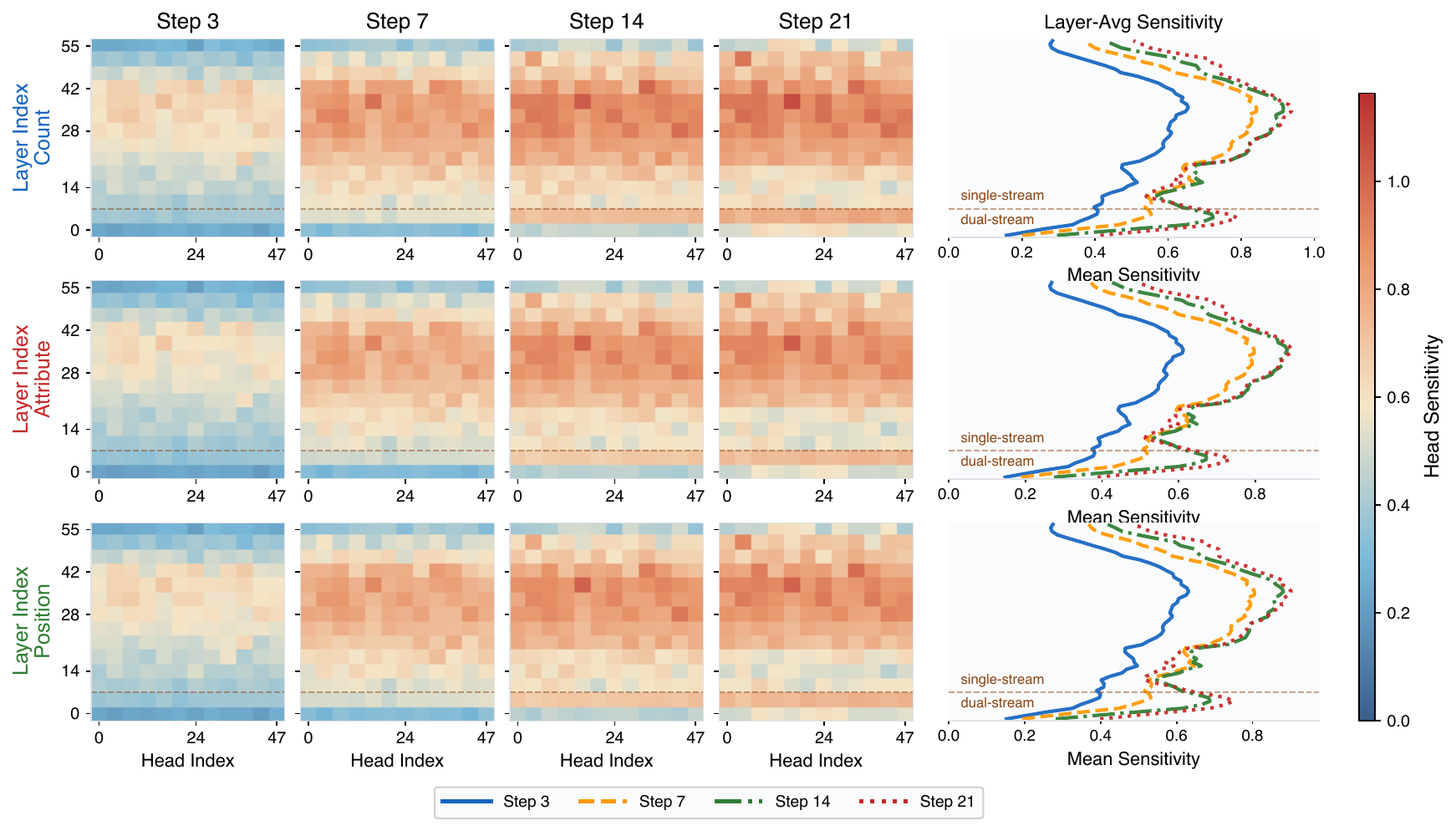}
    \caption{Attention-head sensitivity across denoising steps. Each heatmap cell averages a $4\!\times\!4$ group of transformer layers and attention heads. The side panels show the layer-wise mean sensitivity at each step, and the dashed line separates the dual-stream layers (0--7) from the single-stream layers (8--55).}
    \label{fig:head_sensitivity}
\end{figure}

Figure~\ref{fig:head_sensitivity} shows different temporal patterns across the three skills.
Spatial relation is most sensitive at the early step and then declines; attribute becomes broadly sensitive from step 7 onward; and count is most clearly separated at step 7.
The most sensitive heads also overlap substantially across skills, indicating temporal differences rather than a one-to-one assignment between skills and attention heads.

\section{Evaluation Protocol}\label{app:evaluation_protocol}

\paragraph{Aggregation.}
For GenEval~2, each replicate first averages the Overall score and each skill score over all 800 prompts.
For Qwen-Image-Bench, each replicate first averages the Overall score over all 1,000 prompts.
The Hard Case subset is aggregated the same way: each replicate first averages the GenEval~2 Overall score over the 50 subset prompts.
We then report the mean and sample standard deviation across the four replicates, using $ddof=1$.

\paragraph{Hard Case subset construction.}
Averaged over the four evaluation seeds, the per-prompt GenEval~2 Overall score of the untrained base models is $\geq 0.95$ on 150 of 800 prompts (18.8\%) for FLUX.2-dev and 93 of 800 (11.6\%) for Qwen-Image-2512.
To construct the subset, the frozen FLUX.2-dev base model generates 64 images for each of the 800 evaluation prompts, and the frozen verifier scores them under the GenEval~2 protocol; these scores serve for subset selection only.
We first retain the 100 prompts with the highest within-prompt score variance, which discards both saturated prompts (consistently correct) and uninformative ones (consistently incorrect), and then keep the 50 with the lowest mean score among them.
The subset is fixed before training, involves no trained method, and is shared by all methods on both backbones.

\end{document}